\documentclass{article} 
\pdfoutput=1
\usepackage{nips13submit_e,times}
\usepackage{amsmath}
\usepackage{amssymb}
\usepackage{pgfplots}
\usepackage{pgfplotstable}
\usepackage{tikz}
\usepackage{url}
\usepackage{subcaption}
\pgfplotsset{compat=1.3}

\definecolor{darkblue}{RGB}{47,1,154}
\definecolor{darkred}{RGB}{240,40,40}
\definecolor{darkgreen}{rgb}{0,.6,0}

\title{Improving Deep Neural Networks with
  Probabilistic Maxout Units}

 \author{
 Jost Tobias Springenberg and Martin Riedmiller\\
 Department of Computer Science\\
 University of Freiburg \\
 79110, Freiburg im Breisgau, Germany \\
 \texttt{\{springj,riedmiller\}@cs.uni-freiburg.de} \\
 }

\nipsfinalcopy 

\begin{document}

\maketitle

\begin{abstract}
  We present a probabilistic variant of the recently introduced
  \emph{maxout} unit. The success of deep neural networks utilizing
  \emph{maxout} can partly be attributed to favorable performance
  under dropout, when compared to rectified linear units. It however
  also depends on the fact that each maxout unit performs a pooling
  operation over a group of linear transformations and is thus
  partially invariant to changes in its input. Starting from this
  observation we ask the question: Can the desirable properties of
  \emph{maxout} units be preserved while improving their invariance
  properties ? We argue that our probabilistic maxout (\emph{probout})
  units successfully achieve this balance. We quantitatively verify
  this claim and report classification performance matching or
  exceeding the current state of the art on three challenging image
  classification benchmarks (CIFAR-10, CIFAR-100 and SVHN).
\end{abstract}

\section{Introduction}
Regularization of large neural networks through stochastic model
averaging was recently shown to be an effective tool against
overfitting in supervised classification tasks. Dropout
\cite{Hinton2012} was the first of these stochastic methods which led
to improved performance on several benchmarks ranging from small to
large scale classification
problems~\cite{Krizhevsky2012,Hinton2012}. The idea behind
dropout is to randomly drop the activation of each unit within the
network with a probability of $50 \%$. This can be seen as an extreme
form of bagging in which parameters are shared among models, and the
number of trained models is exponential in the number of these model
parameters. During testing an approximation is used to average over
this large number of models without instantiating each of them.  When
combined with efficient parallel implementations this procedure opened
the possibility to train large neural networks with millions of
parameters via back-propagation \cite{Krizhevsky2012, Zeiler2013}~.

Inspired by this success a number of other stochastic regularization
techniques were recently developed. This includes the work on
dropconnect\cite{WanLi2013}, a generalization of dropout, in which
connections between units rather than their activation are dropped at
random. Adaptive dropout \cite{Frey2013} is a recently introduced
variant of dropout in which the stochastic regularization is performed
through a binary belief network that is learned alongside the neural
network to decrease the information content of its hidden units.
Stochastic pooling \cite{ZeilerStochastic2013} is a technique
applicable to convolutional networks in which the pooling operation is
replaced with a sampling procedure.

Instead of changing the regularizer the authors in
\cite{Goodfellow2013} searched for an activation function for which
dropout performs well. As a result they introduced the maxout unit,
which can be seen as a generalization of rectified linear units
(ReLUs) \cite{Nair2010,Glorot2011}, that is especially suited for the
model averaging performed by dropout. The success of maxout can partly
be attributed to the fact that maxout aids the optimization procedure
by partially preventing units from becoming inactive; an artifact
caused by the thresholding performed by the rectified linear
unit. Additionally, similar to ReLUs, they are piecewise linear
and -- in contrast to e.g. sigmoid units -- typically do not saturate,
which makes networks containing maxout units easier to optimize.

We argue that an equally important property of the maxout unit however
is that its activation function can be seen as performing a pooling
operation over a subspace of $k$ linear feature mappings (in the
following referred to as subspace pooling). As a result of this
subspace pooling operation each maxout unit is partially invariant to
changes within its input. A natural question arising from this
observation is thus whether it could be beneficial to replace the
maximum operation used in maxout units with other pooling operations,
such as L2 pooling. The utility of different subspace pooling
operations has already been explored in the context of unsupervised
learning where e.g. L2-pooling is known give rise to interesting
invariances \cite{Hyvarinnen2009,Bergstra2009,Zou2012}. While work on
generalizing maxout by replacing the max-operation with general
$Lp$-pooling exists \cite{Gulcere2013}, a deviation from the standard
maximum operation comes at the price of discarding some of the
desirable properties of the maxout unit. For example abandoning piecewise
linearity, restricting units to positive values and the introduction
of saturation regimes, which potentially worsen the accuracy of the
approximate model averaging performed by dropout.

Based on these observations we propose a stochastic generalization of
the maxout unit that preserves its desirable properties while
improving the subspace pooling operation of each unit. As an
additional benefit when training a neural network using our proposed
probabilistic maxout units the gradient of the training error is more
evenly distributed among the linear feature mappings of each unit. In
contrast, a maxout network helps gradient flow through each of the
maxout units but not through their k linear feature mappings.
Compared to maxout our probabilistic units thus learn to better
utilize their full k-dimensional subspace. We evaluate the
classification performance of a model consisting of these units and
show that it matches the state of the art performance on three
challenging classification benchmarks.

\section{Model Description}
Before defining the probabilistic maxout unit we briefly review the
notation used in the following for defining deep neural network
models. We adopt the standard feed-forward neural network formulation
in which given an input $\mathbf{x}$ and desired output $y$ (a
class label) the network realizes a function computing a
$C$-dimensional vector $\mathbf{o}$ -- where $C$ is the number of
classes -- predicting the desired output.  The prediction is computed
by first sequentially mapping the input to a hierarchy of $N$ hidden
layers $\mathbf{h}^{(1)}, \dots, \mathbf{h}^{(N)}$. Each unit
$h_i^{(l)}$ within hidden layer $l \in [1,N]$ in the hierarchy
realizes a function $h_i^{(l)}(\mathbf{v}; \mathbf{w}^{(l)}_i,
b^{(l)}_i)$ mapping its inputs $\mathbf{v}$ (given either as the input
$\mathbf{x}$ or the output of the previous layer $h^{(l-1)}$) to an
activation using weight and bias parameters $\mathbf{w}^{(l)}_i$ and
$b^{(l)}_i$. Finally the prediction is computed based on the last
layer output $\mathbf{h}^{N}$. This prediction is realized using a
softmax layer $\mathbf{o} = softmax(\mathbf{W}^{N+1} \mathbf{h}^{(N)}
+ \mathbf{b}^{N+1})$ with weights $\mathbf{W}^{N+1}$ and bias
$\mathbf{b}^{N+1}$. All parameters $\theta = \{ W^{(1)}, b^{(1)},
\dots, W^{(N+1)}, b^{(N+1)} \}$ are then learned by minimizing the
cross entropy loss between output probabilities $\mathbf{o}$ and label
$y$ : $ \mathcal{L}(o, y; \mathbf{x}) = - \sum_{i = 1}^C y_i \log(o_i)
+ (1 - y_i) log(1 - o_i)$.

\subsection{Probabilistic Maxout Units}
\begin{figure}
\centering
\includegraphics[width=\columnwidth]{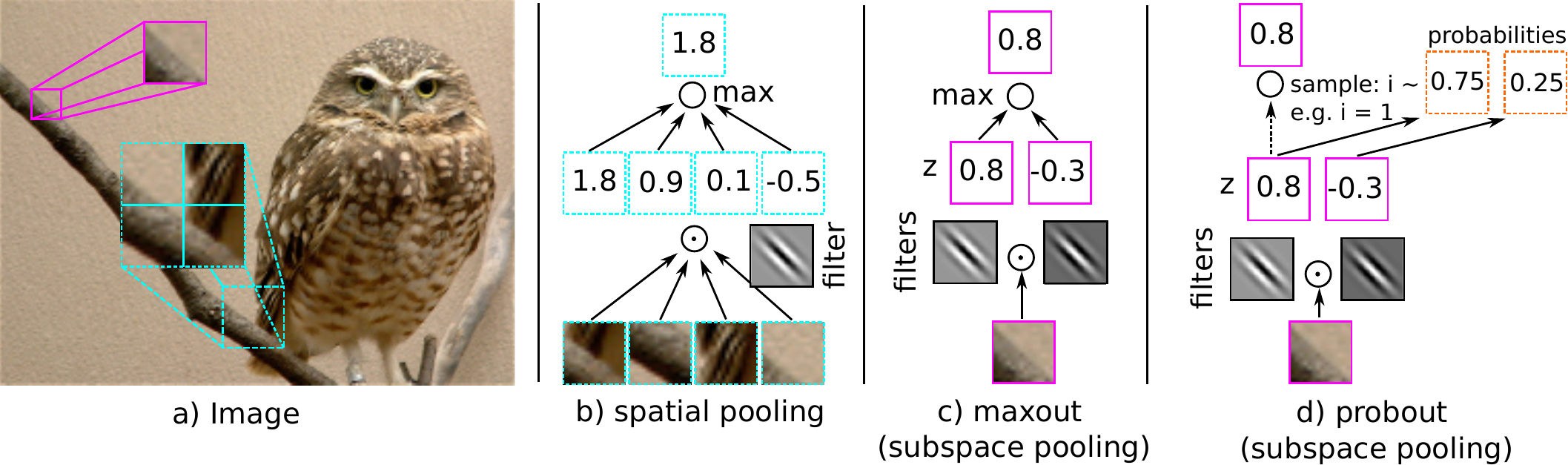}
\caption{Schematic of different pooling
  operations. a) An exemplary input image taken from the ImageNet
  dataset together with the depiction of a spatial pooling region (cyan) as well as
  the input to one maxout / probout unit (marked in magenta). b)
  Spatial max-pooling proceeds by computing the maximum of one filter
  response at the four different positions from a). c) Maxout computes
  a pooled response of two linear filter mappings applied to
  one input patch. d) The activation of a probout unit is computed by
  sampling one of the linear responses according to their probability.}
\label{pooling_fig}
\end{figure}

The maxout unit was recently introduced in \cite{Goodfellow2013} and can be
formalized as follows: Given the units input $\mathbf{v} \in
\mathbb{R}^d$ (either the activation from the previous
layer or the input vector) the activation of a maxout unit is
computed by first computing k linear
feature mappings $\mathbf{z} \in
\mathbb{R}^{k}$ where
\begin{equation}
  z_i = \mathbf{w}_i \mathbf{v} + b_i,
\end{equation}
and k is the number of linear sub-units combined by one maxout unit.
Afterwards the output $h_{maxout}$ of the maxout hidden unit is given
as the maximum over the k feature mappings:
\begin{equation}
  h_{maxout}(\mathbf{v}) = \max [z_1, \dots, z_k].
  \label{maxout_act}
\end{equation}
When formalized like this it becomes clear that (in contrast to
conventional activation functions) the maxout unit can be interpreted
as performing a pooling operation over a k-dimensional subspace of
linear units $[z_1, \dots, z_k]$ each representing one transformation
of the input $\mathbf{v}$. This is similar to spatial max-pooling
which is commonly employed in convolutional neural networks. However,
unlike in spatial pooling the maxout unit pools over a subspace of k
different linear transformations applied to the same input
$\mathbf{v}$. In contrast to this, spatial max-pooling of linear
feature maps would compute a pooling over one linear transformation
applied to k different inputs. A schematic of the difference between
several pooling operations is given in Fig. \ref{pooling_fig}~.

As such maxout is thus more similar to the subspace pooling operations
used for example in topographic ICA \cite{Hyvarinnen2009} which is
known to result in partial invariance to changes within its input. On the
basis of this observation we propose a stochastic generalization of
the maxout unit that preserves its desirable properties while
improving gradient propagation among the $k$ linear feature mappings
as well as the invariance properties of each unit. In the following we
call these generalized units \emph{probout units} since they are a
direct probabilistic generalization of maxout.

We derive the probout unit activation function from the maxout
formulation by replacing the maximum operation in
Eq. \eqref{maxout_act} with a probabilistic sampling procedure. More
specifically we assume a Boltzmann distribution over the $k$ linear
feature mappings and sample the
activation $h(\mathbf{v})$ from the activation of the corresponding
subspace units. To this end we first define a probability for each of
the k linear units in the subspace as:
\begin{equation}
  p_i = \frac{e^{\lambda z_i}}{\sum_{j=1}^k e^{\lambda z_j}},
\end{equation} 
where $\lambda$ is a hyperparameter (referred to as an inverse
temperature parameter) controlling the variance of the distribution.
The activation $h_{probout}(\mathbf{x})$ is then sampled as 
\begin{equation}
  h_{probout}(\mathbf{v}) = z_i, \text{ where } i \sim Multinomial\{p_1, \dots, p_k\}.
  \label{probout_act}
\end{equation}  
Comparing Eq. \eqref{probout_act} to Eq. \eqref{maxout_act} we see
that both, are not bounded from above or below and their activation is
always given as one of the linear feature mappings within their
subspace. The probout unit hence preserves most of the properties of
the maxout unit, only replacing the sub-unit selection mechanism.

We can further see that Eq. \eqref{probout_act} reduces to the maxout
activation for $\lambda \rightarrow \infty$. For other values of
$\lambda$ the probout unit will behave similarly to maxout when the
activation of one linear unit in the subspace dominates. However, if
the activation of multiple linear units differs only slightly they will
be selected with almost equal probability. Futhermore, each active
linear unit will have a chance to be selected. The sampling approach
therefore ensures that gradient flows through each of the $k$ linear
subspace units of a given probout unit for some examples (given that
$\lambda$ is sufficiently small). We hence argue that probout units can learn to better
utilize their full k-dimensional subspace. 

In practice we want to combine the probout units described by
Eq. \eqref{probout_act} with dropout for regularizing the learned
model.  To achieve this we directly include dropout in the
probabilistic sampling step by re-defining the probabilities as:
\begin{align}
  \hat{p}_0 &= 0.5 \\
  \hat{p}_i &= \frac{e^{\lambda z_i}}{2 \cdot \sum_{j=1}^k e^{\lambda z_j}}.
\end{align}
Consequently, we sample the probout activation function
including dropout $\hat{h}_{probout}(\mathbf{v})$ as
\begin{equation}
  \hat{h}_{probout}(\mathbf{v}) = \begin{cases} 
    0 \text{ if } i = 0 \\
    z_i \text{ else }
  \end{cases}, \text{ where } i \sim Multinomial\{\hat{p}_0 , \hat{p}_1, \dots,
  \hat{p}_k \}.
  \label{probout_actdo}
\end{equation}

\subsection{Relation to other pooling operations}
The idea of using a stochastic pooling operation has been explored in
the context of spatial pooling within the machine learning literature
before. Among this work the approach most similar to ours is
\cite{Lee2009}. There the authors introduced a probabilistic
pooling approach in order to derive a convolutional deep believe
network (DBN). They also use a Boltzmann distribution based on unit
activations to calculate a sampling probability. The main difference
between their work and ours is that they calculate the probability of
sampling one unit at different spatial locations whereas we calculate
the probability of sampling a unit among k units forming a subspace at
one spatial location.
Another difference is that we forward propagate the sampled activation
$z_i$ whereas they use the calculated probability to activate a binary
stochastic unit.

Another approach closely related to our work is the stochastic pooling
presented in \cite{ZeilerStochastic2013}. Their stochastic pooling
operation samples the activation of a pooling unit $p_i$
proportionally to the activation $a$ of a rectified linear unit
\cite{Nair2010} computed at different spatial positions. This is
similar to Eq. \eqref{probout_act} in the sense that the activation is
sampled from a set of different activations. Similar to
\cite{Lee2009} it however differs in that the sampling is
performed over spatial locations rather than activations of different
units.

It should be noted that our work also bears some resemblance to recent
work on training stochastic units, embedded in an autoencoder network,
via back-propagation \cite{BengioStochastic2013,BengioGSN2013}. In contrast
to their work, which aims at using stochastic neurons to train a
generative model, we embrace stochasticity in the subspace pooling
operation as an effective means to regularize a discriminative
model.

\subsection{Inference}
\label{sect_inference}
At test time we need to account for the stochastic nature of a neural
network containing probout units. During a forward pass through the
network the value of each probout unit is sampled from one of $k$
values according to their probability. The output of such a forward
pass thus always represents only one of $k^M$ different instantiations
of the trained probout network; where $M$ is the number of probout
units in the network. When combined with dropout the number of
possible instantiations increases to ${(k+1)}^M$. Evaluating all
possible models at test time is therefore clearly infeasible. The
Dropout formulation from \cite{Hinton2012} deals with this large amount of
possible models by removing dropout at test time and halving the
weights of each unit. If the network consists of only one softmax layer
then this modified network performs exact model averaging
\cite{Hinton2012}.  For general models this computation is merely
an approximation of the true model average which, however, performs
well in practice for both deep ReLU networks \cite{Krizhevsky2012} and
the maxout model \cite{Goodfellow2013}.

We adopt the same procedure of halving the weights for removing the
influence of dropout at test-time and rescale the probabilities such
that $\sum_{i=1}^k \hat{p}_i = 1$ and $\hat{p}_0 = 0$, effectively replacing the
sampling from Eq .\eqref{probout_actdo} with
Eq. \eqref{probout_act}. We further observe that from the $k^M$ models
remaining after removing dropout only few models will be instantiated
with high probability. We therefore resort to sampling a small number
of outputs $\mathbf{o}$ from the networks softmax layer and average
their values. An evaluation of the exact effect of this
model averaging can be found in Section \ref{sect_prelim_eval}~.

\section{Evaluation}
We evaluate our method on three different image classification
datasets (CIFAR-10, CIFAR-100 and SVHN) comparing it against the basic
maxout model as well as the current state of the art on all
datasets. All experiments were performed using an implementation based
on Theano and the pylearn2 library \cite{GoodfellowPylearn} using the
fast convoltion code of \cite{Krizhevsky2012}. We use mini-batch
stochastic gradient descent with a batch size of 100. For each of the
datasets we start with the same network used in
\cite{Goodfellow2013}~-- retaining all of their hyperparameter
choices -- to ensure comparability between results. We replace the
maxout units in the network with probout units and choose one
$\lambda^{(l)}$ via crossvalidation for each layer $l$ in a preliminary
experiment on CIFAR-10.

\subsection{Experiments on CIFAR-10}
We begin our experiments with the CIFAR-10 \cite{Krizhevsky2009}
dataset. It consists of $50,000$ training images and $10,000$ test
images that are grouped into $10$ categories. Each of these images is
of size $32\times32$ pixels and contains $3$ color channels. Maxout is
known to yield good performance on this dataset, making it an ideal
starting point for evaluating the difference between maxout and
probout units.

\subsubsection{Effect of replacing maxout with probout units}
\label{sect_prelim_eval}
\begin{figure}
\centering
\includegraphics{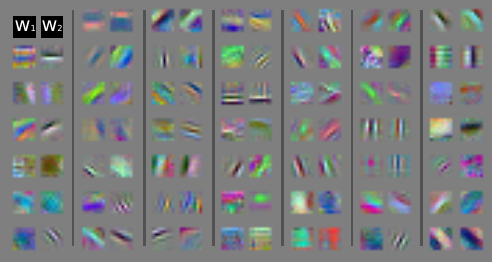}
\quad
\includegraphics{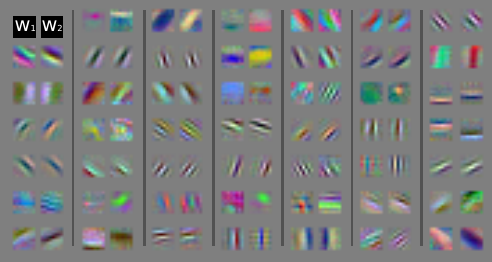}
\caption{Visualization of pairs of first layer linear filters learned by the maxout
  model (left) as well as the probout model (right). In contrast to
  the maxout filters the filter pairs learned by the probout model
  appear to mostly be transformed versions of each other.}
\label{filters_fig}
\end{figure}
We conducted a preliminary experiment to evaluate the effect of the
probout parameters $\lambda^{(l)}$ on the performance and compare it to the
standard maxout model. For this purpose we use a five layer model consisting of
three convolutional layers with 48, 128 and 128 probout units
respectively which pool over 2 linear units each. The penultimate
layer then consists of 240 probout units pooling over a subspace of 5
linear units. The final layer is a standard softmax layer mapping from
the 240 units in the penultimate layer to the 10 classes of CIFAR-10.
The receptive fields of units in the convolutional layers are 8, 8 and
5 respectively. Additionally, spatial max-pooling is performed after
each convolutional layer with pooling size of $4 \times 4$, $4 \times
4$ and $2 \times 2$ using a stride of 2 in all layers.
We split the CIFAR-10 training data retaining the first 40000
samples for training and using the last 10000 samples as a validation
set.

We start our evaluation by using probout units everywhere in the
network and cross-validate the choice of the inverse-temperature
parameters $\lambda^{(l)} \in \{ 0.1, 0.5, 1, 2, 3, 4 \}$ keeping all
other hyperparameters fixed. We find that annealing the
$\lambda^{(l)}$ parameter during training to a lower value improved
performance for all $\lambda^{(l)} > 0.5$ and hence linearly decrease
$\lambda^{(l)}$ to a value that is $0.9$ lower than the initial
$\lambda$ in these cases. As shown in Fig. \ref{lambda_plot} the best
classification performance is achieved when $\lambda$ is set to allow
higher variance sampling for the first two layers, specifically when
$\lambda^{(1)} = 1$ and $\lambda^{(2)} = 2$. For the third as well as
the fully connected layer we observe a performance increase when
$\lambda^{(3)}$ is chosen as $\lambda^{(3)} = 3$ and $\lambda^{(4)} =
4$, meaning that the sampling procedure selects the maximum value with
high probability. This indicates that the probabilistic sampling is
most effective in lower layers.  We verified this by replacing the
probout units in the last two layers with maxout units which did not
significantly decrease classification accuracy.

\begin{figure}[h]
\centering
\begin{subfigure}[b]{0.48\columnwidth}
  \begin{tikzpicture}
    \begin{axis}[
      compat=1.3,
      width=\columnwidth, height=5.5cm,     
      grid = major,
      grid style={dashed, gray!30},
      xmin=0.1,     
      xmax=4,    
      ymin=12.5,     
      ymax=13.5,   
      xtick={0.1,0.5,1,2,3,4},
      ytick={12.5,12.6,...,13.5},
      legend entries={
        {$l = 1$ (with $\lambda^{(2)} = 2$)},
        {$l = 2$ (with $\lambda^{(1)} = 1$)}
      },
      legend reversed,
      legend pos=north east,
      legend cell align=left,
      legend style={fill = none,font=\small},
      axis background/.style={fill=white},
      ylabel={Class. Error (validation set)},
      xlabel=$\lambda^{(l)}$,
      tick align=outside]
      
      \addplot+[error bars/.cd,y dir=both] 
      table[x=lambda,y=lambda1] {lamdadata.dat};

      \addplot+[error bars/.cd,y dir=both] 
      table[x=lambda,y=lambda2] {lamdadata.dat};

    \end{axis}
  \end{tikzpicture}
  \vspace{-20px}
  \caption{}
  \label{lambda_plot}
\end{subfigure}
\begin{subfigure}[b]{0.5\columnwidth}
  \begin{tikzpicture}
    \begin{axis}[
      compat=1.3,
      width=\columnwidth, height=5cm,     
      grid = major,
      grid style={dashed, gray!30},
      xmin=0,     
      xmax=100,    
      ymin=12.3,     
      ymax=16.3,   
      xtick={1,5,10,20,30,...,100},
      ytick={10.0,11.0,...,15.0},
      legend entries={
        Maxout + sampling,
        Probout,
        Maxout baseline
      },
      legend reversed,
      legend style={at={(0.97,0.5)},anchor=east},
      legend cell align=left,
      legend style={fill = none,font=\small},
      axis background/.style={fill=white},
      ylabel=Classification error,
      xlabel=Model evaluations,
      tick align=outside]
      
      \addplot+[error bars/.cd,y dir=both, y explicit] 
      table[x=X,y=Maxout,y error=MaxoutSTD] {dataaveraging.dat};

      \addplot+[error bars/.cd,y dir=both, y explicit] 
      table[x=X,y=Probout,y error=ProboutSTD]
      {dataaveraging.dat};

      \addplot[domain=0:100, green, thick, dashed] 
      {13.10}; 

    \end{axis}
  \end{tikzpicture}
  \caption{}
  \label{model_average_plot}
\end{subfigure}
\caption{(a) Validation of the $\lambda^{(l)}$ parameter for layers $l
  \in [1, 2]$ on CIFAR-10. We plot the error on the validation set
  after training (using 50 model evaluations). When evaluating the
  choice of $\lambda^{(1)}$ (red curve) the second parameter fixed
  $\lambda^{(2)} = 2$. Likewise, for the experiments regarding
  $\lambda^{(2)}$ (blue curve) $\lambda^{(1)} = 1$.  (b) Evolution of
  the classification error and standard deviation on the CIFAR-10
  dataset for a changing number $E$ of model evaluations. We average
  the activation $\mathbf{o} \in \mathbb{R}^{C}$ of the softmax layer
  over all $E$ evaluations and compute the predicted class label
  $\hat{y}$ as the maximum $\hat{y} = \arg \max_{i \in \{ 1, \dots, C
    \}} o_i$. The standard deviation is computed over 10 runs of $E$
  model evaluations.}
  \label{two_plots}
\end{figure}

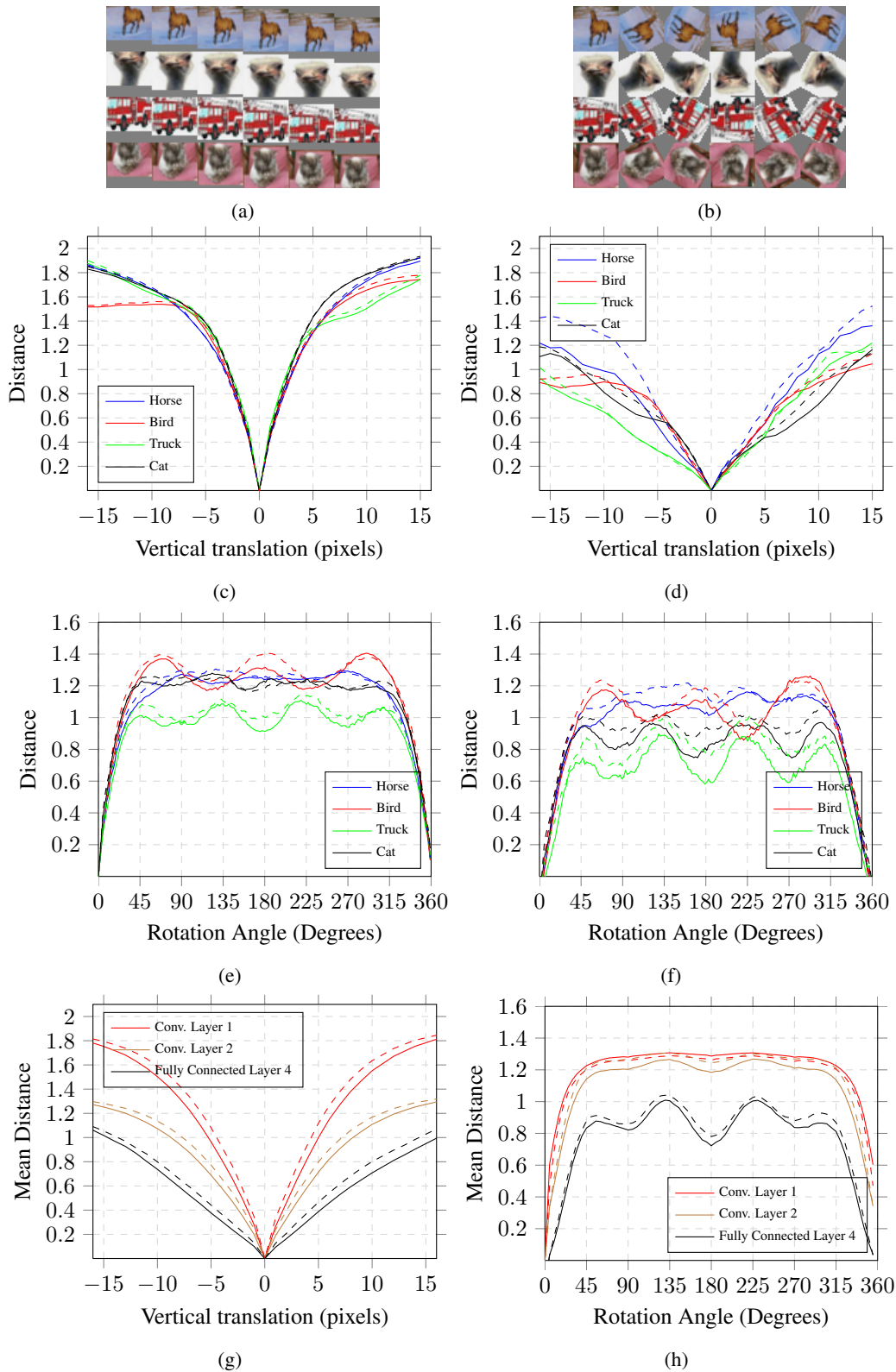
\begin{figure}[h]
\vspace{-50px}
\hspace{20px}
\begin{subfigure}[b]{0.3\columnwidth}
\includegraphics[width=\columnwidth]{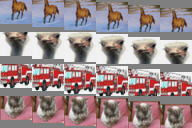}
  \caption{}
  \label{images_translation}
\end{subfigure}
\hspace{80px}
\begin{subfigure}[b]{0.3\columnwidth}
\includegraphics[width=\columnwidth]{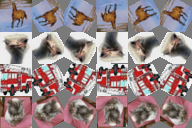}
  \caption{}
  \label{images_rotation}
\end{subfigure}
\centering
\begin{subfigure}[b]{0.492\columnwidth}
  \begin{tikzpicture}
    \begin{axis}[
      compat=1.3,
      width=\columnwidth, height=5.5cm,     
      grid = major,
      grid style={dashed, gray!30},
      xmin=-16.,     
      xmax=16,    
      ymin=0.,     
      ymax=2.1,   
      xtick={-15,-10,-5,...,15},
      ytick={0.2,0.4,...,2.1},
      legend entries={
        {Horse},
        {Bird},
        {Truck},
        {Cat}
      },
      legend pos=south west,
      legend cell align=left,
      legend style={fill = none,font=\tiny},
      axis background/.style={fill=white},
      ylabel={Distance},
      xlabel={Vertical translation (pixels)},
      tick align=outside]
      
      \pgfplotsset{table/col sep = comma}

      \addplot+[mark=none, blue] 
      table[x=px,y=conv1] {output_probout_trans_single_10_new.val};
      \addplot+[mark=none, red] 
      table[x=px,y=conv1] {output_probout_trans_single_20_new.val};
      \addplot+[mark=none, green] 
      table[x=px,y=conv1] {output_probout_trans_single_40_new.val};
      \addplot+[mark=none, black] 
      table[x=px,y=conv1] {output_probout_trans_single_66_new.val};
      
      \addplot+[mark=none, blue, dashed, forget plot] 
      table[x=px,y=conv1] {output_maxout_trans_single_10_new.val};
      \addplot+[mark=none, red, dashed, forget plot] 
      table[x=px,y=conv1] {output_maxout_trans_single_20_new.val};
      \addplot+[mark=none, green, dashed, forget plot] 
      table[x=px,y=conv1] {output_maxout_trans_single_40_new.val};
      \addplot+[mark=none, black, dashed, forget plot] 
      table[x=px,y=conv1] {output_maxout_trans_single_66_new.val};
    \end{axis}
  \end{tikzpicture}
  \caption{}
  \label{translation_single_plot}
\end{subfigure}
\begin{subfigure}[b]{0.492\columnwidth}
  \begin{tikzpicture}
    \begin{axis}[
      compat=1.3,
      width=\columnwidth, height=5.5cm,     
      grid = major,
      grid style={dashed, gray!30},
      xmin=-16.,     
      xmax=16,    
      ymin=0.,     
      ymax=2.1,   
      xtick={-15,-10,-5,...,15},
      ytick={0.2,0.4,...,2.1},
      legend entries={
        {Horse},
        {Bird},
        {Truck},
        {Cat}
      },
      legend pos=north west,
      legend cell align=left,
      legend style={fill = none,font=\tiny},
      axis background/.style={fill=white},
      ylabel={Distance},
      xlabel={Vertical translation (pixels)},
      tick align=outside]
      
      \pgfplotsset{table/col sep = comma}
      \addplot+[mark=none, blue] 
      table[x=px,y=fcon] {output_probout_trans_single_10_new.val};
      \addplot+[mark=none, red] 
      table[x=px,y=fcon] {output_probout_trans_single_20_new.val};
      \addplot+[mark=none, green] 
      table[x=px,y=fcon] {output_probout_trans_single_40_new.val};
      \addplot+[mark=none, black] 
      table[x=px,y=fcon] {output_probout_trans_single_66_new.val};

      \addplot+[mark=none, blue, dashed, forget plot] 
      table[x=px,y=fcon] {output_maxout_trans_single_10_new.val};
      \addplot+[mark=none, red, dashed, forget plot] 
      table[x=px,y=fcon] {output_maxout_trans_single_20_new.val};
      \addplot+[mark=none, green, dashed, forget plot] 
      table[x=px,y=fcon] {output_maxout_trans_single_40_new.val};
      \addplot+[mark=none, black, dashed, forget plot] 
      table[x=px,y=fcon] {output_maxout_trans_single_66_new.val};

    \end{axis}
  \end{tikzpicture}
  \caption{}
  \label{translation_single_plot_last}
\end{subfigure}
\begin{subfigure}[b]{0.48\columnwidth}
  \begin{tikzpicture}
    \begin{axis}[
      compat=1.3,
      width=\columnwidth, height=5.5cm,     
      grid = major,
      grid style={dashed, gray!30},
      xmin=0,     
      xmax=360,    
      ymin=0.,     
      ymax=1.6,   
      xtick={0,45,...,360},
      ytick={0.2,0.4,...,1.6},
      legend entries={
        {Horse},
        {Bird},
        {Truck},
        {Cat}
      },
      legend pos=south east,
      legend cell align=left,
      legend style={fill = none,font=\tiny},
      axis background/.style={fill=white},
      ylabel={Distance},
      xlabel={Rotation Angle (Degrees)},
      tick align=outside]
      
      \pgfplotsset{table/col sep = comma}

      \addplot+[mark=none, blue] 
      table[x=px,y=conv2] {output_probout_rot_single_10_new.val};
      \addplot+[mark=none, red] 
      table[x=px,y=conv2] {output_probout_rot_single_20_new.val};
      \addplot+[mark=none, green] 
      table[x=px,y=conv2] {output_probout_rot_single_40_new.val};
      \addplot+[mark=none, black] 
      table[x=px,y=conv2] {output_probout_rot_single_66_new.val};

      \addplot+[mark=none, blue, dashed, forget plot] 
      table[x=px,y=conv2] {output_maxout_rot_single_10_new.val};
      \addplot+[mark=none, red, dashed, forget plot] 
      table[x=px,y=conv2] {output_maxout_rot_single_20_new.val};
      \addplot+[mark=none, green, dashed, forget plot] 
      table[x=px,y=conv2] {output_maxout_rot_single_40_new.val};
      \addplot+[mark=none, black, dashed, forget plot] 
      table[x=px,y=conv2] {output_maxout_rot_single_66_new.val};
    \end{axis}    
  \end{tikzpicture}
  \caption{}
  \label{rotation_single_plot}
\end{subfigure}
\begin{subfigure}[b]{0.48\columnwidth}
  \begin{tikzpicture}
    \begin{axis}[
      compat=1.3,
      width=\columnwidth, height=5.5cm,     
      grid = major,
      grid style={dashed, gray!30},
      xmin=0,     
      xmax=360,    
      ymin=0.,     
      ymax=1.6,   
      xtick={0,45,...,360},
      ytick={0.2,0.4,...,1.6},
      legend entries={
        {Horse},
        {Bird},
        {Truck},
        {Cat}
      },
      legend pos=south east,
      legend cell align=left,
      legend style={fill = none,font=\tiny},
      axis background/.style={fill=white},
      ylabel={Distance},
      xlabel={Rotation Angle (Degrees)},
      tick align=outside]
      
      \pgfplotsset{table/col sep = comma}

      \addplot+[mark=none, blue] 
      table[x=px,y=conv3] {output_probout_rot_single_10_new.val};
      \addplot+[mark=none, red] 
      table[x=px,y=conv3] {output_probout_rot_single_20_new.val};
      \addplot+[mark=none, green] 
      table[x=px,y=conv3] {output_probout_rot_single_40_new.val};
      \addplot+[mark=none, black] 
      table[x=px,y=conv3] {output_probout_rot_single_66_new.val};

      \addplot+[mark=none, blue, dashed, forget plot] 
      table[x=px,y=conv3] {output_maxout_rot_single_10_new.val};
      \addplot+[mark=none, red, dashed, forget plot] 
      table[x=px,y=conv3] {output_maxout_rot_single_20_new.val};
      \addplot+[mark=none, green, dashed, forget plot] 
      table[x=px,y=conv3] {output_maxout_rot_single_40_new.val};
      \addplot+[mark=none, black, dashed, forget plot] 
      table[x=px,y=conv3] {output_maxout_rot_single_66_new.val};
    \end{axis}    
  \end{tikzpicture}
  \caption{}
  \label{rotation_single_plot_last}
\end{subfigure}
\begin{subfigure}[b]{0.492\columnwidth}
  \begin{tikzpicture}
    \begin{axis}[
      compat=1.3,
      width=\columnwidth, height=5.5cm,     
      grid = major,
      grid style={dashed, gray!30},
      xmin=-16.,     
      xmax=16,    
      ymin=0.,     
      ymax=2.1,   
      xtick={-15,-10,-5,...,15},
      ytick={0.2,0.4,...,2.1},
      legend entries={
        {Conv. Layer 1},
        {Conv. Layer 2},
        {Fully Connected Layer 4},
      },
      legend pos=north west,
      legend cell align=left,
      legend style={fill = none,font=\tiny},
      axis background/.style={fill=white},
      ylabel={Mean Distance},
      xlabel={Vertical translation (pixels)},
      tick align=outside]
      
      \pgfplotsset{table/col sep = comma}
      \addplot+[mark=none, red] 
      table[x=px,y=conv1] {output_probout_trans_mean_new.val};
      \addplot+[mark=none, brown] 
      table[x=px,y=conv3] {output_probout_trans_mean_new.val};
      \addplot+[mark=none, black] 
      table[x=px,y=fcon] {output_probout_trans_mean_new.val};
      
      \addplot+[mark=none, red, dashed, forget plot] 
      table[x=px,y=conv1] {output_maxout_trans_mean_new.val};
      \addplot+[mark=none, brown, dashed, forget plot] 
      table[x=px,y=conv3] {output_maxout_trans_mean_new.val};
      \addplot+[mark=none, black, dashed, forget plot] 
      table[x=px,y=fcon] {output_maxout_trans_mean_new.val};
    \end{axis}    
  \end{tikzpicture}
  \caption{}
  \label{translation_single_plot}
\end{subfigure}
\begin{subfigure}[b]{0.48\columnwidth}
  \begin{tikzpicture}
    \begin{axis}[
      compat=1.3,
      width=\columnwidth, height=5.5cm,     
      grid = major,
      grid style={dashed, gray!30},
      xmin=0,     
      xmax=360,    
      ymin=0.,     
      ymax=1.6,   
      xtick={0,45,...,360},
      ytick={0.2,0.4,...,1.6},
      legend entries={
        {Conv. Layer 1},
        {Conv. Layer 2},
        {Fully Connected Layer 4},
      },
      legend pos=south east,
      legend cell align=left,
      legend style={fill = none,font=\tiny},
      axis background/.style={fill=white},
      ylabel={Mean Distance},
      xlabel={Rotation Angle (Degrees)},
      tick align=outside]
      
      \pgfplotsset{table/col sep = comma}
      \addplot+[mark=none, red] 
      table[x=px,y=conv1] {output_probout_rot_mean_new.val};
      \addplot+[mark=none, brown] 
      table[x=px,y=conv3] {output_probout_rot_mean_new.val};
      \addplot+[mark=none, black] 
      table[x=px,y=fcon] {output_probout_rot_mean_new.val};

      \addplot+[mark=none, red, dashed, forget plot] 
      table[x=px,y=conv1] {output_maxout_rot_mean_new.val};
      \addplot+[mark=none, brown, dashed, forget plot] 
      table[x=px,y=conv3] {output_maxout_rot_mean_new.val};
      \addplot+[mark=none, black, dashed, forget plot] 
      table[x=px,y=fcon] {output_maxout_rot_mean_new.val};
    \end{axis}    
  \end{tikzpicture}
  \caption{}
  \label{rotation_single_plot}
\end{subfigure}
\caption{Analysis of the impact of vertical translation and rotation
  on features extracted from a maxout and probout network. We plot the
  distance between normalized feature vectors extracted on transformed
  images and the original, unchanged, image. The distances for the
  probout model are plotted using thick lines. The distances for the
  maxout model are depicted using dashed lines. (a,b) 4 exemplary images
  undergoing different vertical translations and rotations
  respectively. (c,d) Euclidean distance between feature vectors from
  the original 4 images depicted in (a,b) and transformed images for
  Layer 1 (convolutional) and Layer 4 (fully connected) respectively. (e,f) Euclidean distance between feature vectors from
  the original 4 images  and transformed versions for Layer 2
  (convolutional) and Layer 4 (fully connected) respectively.
  (g,h) Mean Euclidean distance between feature vectors extracted from
  100 randomly selected images and their transformed versions for
  different layers in the network.}
  \label{plots_invariance}
\end{figure}

We hypothesize that increasing the probability of sampling a non
maximal linear unit in the subspace pulls the units in the subspace
closer together and forces the network to become ``more invariant'' to
changes within this subspace. This is a property that is desired in
lower layers but might turn to be detrimental in higher layers where
the model averaging effect of maxout is more important than achieving
invariance. Here sampling units with non-maximal activation could
result in unwanted correlation between the ``submodels''. To
qualitatively verify this claim we plot the first layer linear filters
learned using probout units alongside the filters learned by a model
consisting only of maxout units in Fig. \ref{filters_fig}.  When
inspecting the filters we can see that many of the filters belonging
to one subspace formed by a probout unit seem to be transformed
versions of each other, with  some of then resembling ``quadrature pairs''
of filters. Among the linear filters learned by the maxout model
some also appear to encode invariance to local transformations. Most
of the filters contained in a subspace however are seemingly
unrelated.
To support this observation empirically we probed for changes in the
feature vectors of different layers (extracted from both maxout and probout models) when
they are applied to translated and rotated images from the validation set. Similar to
\cite{Koray_CVPR2009,Zeiler2013} we calculate the normalized Euclidean distance between feature
vectors extracted from an unchanged image and a transformed
version. We then plot these distances for several exemplary images as
well as the mean over 100 randomly sampled images. The result of this
experiment is given in Fig. \ref{plots_invariance}, showing that
introducing probout units into the network has a moderate positive effect on both
invariance to translation and rotations.

Finally, we evaluate the computational cost of the model averaging
procedure described in Section \ref{sect_inference} at test time. As
depicted in Fig. \ref{model_average_plot} the classification error for
the probout model decreases with more model evaluations saturating
when a moderate amount of 50 evaluations is reached. Conversely, using
sampling at test time in conjunction with the standard maxout model
significantly decreases performance. This indicates that the maxout
model is highly optimized for the maximum responses and cannot deal
with the noise introduced through the sampling procedure.  We
additionally also tried to replace the model averaging mechanism with
cheaper approximations. Replacing the sampling in the probout units
with a maximum operation at test time resulted in a decrease in
performance, reaching $14.13 \%$.  We also tried to use probability
weighting during testing \cite{ZeilerStochastic2013} which however
performed even worse, achieving $15.21 \%$.

\subsubsection{Evaluation of Classification Performance}
As the next step, we evaluate the performance of our model on the full
CIFAR-10 benchmark.  We follow the same protocol as in
\cite{Goodfellow2013} to train the probout model. That is, we first
preprocess all images by applying contrast normalization followed by
ZCA whitening. We then train our model using the first 40000 examples
from the training set using the last 10000 examples as a validation
set. Training then proceeds until the validation error stops
decreasing. We then retrain the model on the complete training set for
the same amount of epochs it took to reach the best validation error.

\begin{table}[t]
\vskip 0.15in
\caption{Classification error of different models on the
  CIFAR-10 dataset.}
\begin{center}
\begin{small}
\begin{sc}
\begin{tabular}{l|l}
Method  & Error \\
\hline
Conv. Net + Spearmint \cite{Snoek2012} & 14.98 $\%$ \\
Conv. Net + Maxout    \cite{Goodfellow2013}  & 11.69 $\%$ \\
Conv. Net + Probout   & \textbf{11.35 $\%$} \\
\hline
12 $\times$ Conv. Net + dropconnect \cite{WanLi2013} & \textbf{9.32 $\%$} \\
Conv. Net + Maxout \cite{Goodfellow2013} &
                                                                    9.38 $\%$ \\
Conv. Net + Probout & 9.39 $\%$ \\
\hline
\end{tabular}
\end{sc}
\end{small}
\end{center}
\vskip -0.1in
\label{cifar10_results}
\end{table}
To comply with the experiments in \cite{Goodfellow2013} we used a
larger version of the model from Section \ref{sect_prelim_eval} in all
experiments. Compared to the preliminary experiment the size
of the convolutional layers was increased to $96$, $192$ and $192$ units
respectively. The size of the fully connected layer was increased to
500 probout units pooling over a 5 dimensional subspace.

The top half of Table \ref{cifar10_results} shows the result of
training this model as well as other recent results. We achieve an
error of $11.35 \%$, slightly better than -- but statistically tied to --
the previous state of the art given by the maxout model. We also evaluated the performance of this
model when the training data is augmented with additional
transformed training examples. For this purpose we train our model
using the original training images as well as add randomly translated
and horizontally flipped versions of the images. The bottom half of
Table \ref{cifar10_results} shows a comparison of different results
for training on CIFAR-10 with additional data augmentation. Using this
augmentation process we achieve a classification error of $9.39 \%$,
matching, but not outperforming the maxout result.

\subsection{CIFAR-100}
The images contained in the CIFAR-100 dataset \cite{Krizhevsky2009}
are -- just as the CIFAR-10 images -- taken from a subset of the
10-million images database. The dataset contains $50,000$ training and
$10,000$ test examples of size $32\times32$ pixels each. The dataset
is hence similar to CIFAR-10 in both size and image content. It,
however, differs from CIFAR-10 in its label distribution. Concretely,
CIFAR-100 contains images of 100 classes grouped into 20
``super-classes''. The training data therefore contains 500 training
images per class -- 10 times less examples per class than in
CIFAR-10 -- which are accompanied by 100 examples in the test-set.

We do not make use of the 20 super-classes and train a model using a
similar setup to the experiments we carried out on CIFAR-10.
Specifically, we use the same preprocessing and training procedure
(determining the amount of epochs using a validation set and then
retraining the model on the complete data). The same network as in
Section \ref{sect_prelim_eval} was used for this experiment (adapted
to classify 100 classes). Again, this is the same architecture used
in \cite{Goodfellow2013} thus ensuring comparability between results.
During testing we use 50 model evaluations to average over the sampled
\emph{probout} units.

The result of this experiment is given in Table
\ref{cifar100_results}. In agreement with the CIFAR-10 results our
model performs marginally better than the maxout model (by $0.45 \%$\footnote{While we were
  writing this manuscript it came to our attention that the
  experiments on CIFAR-100 in \cite{Goodfellow2013} were carried out
  using a different preprocessing than mentioned in the original paper. To
  ensure that this does not substantially effect our comparison we ran
  their experiment using the same preprocessing used in our
  experiments. This resulted in a slightly improved classification
  error of $38.50 \%$.}). As
also shown in the table the current best method on CIFAR-100 achieves
a classification error of $36.85 \%$ \cite{Nitish2013}, using a larger
convolutional neural network together with a tree-based prior on the
classes formed by utilizing the super-classes. A similar
performance increase could potentially be achieved by combining their
tree-based prior with our model.

\subsection{SVHN}
The street view house numbers dataset \cite{Netzer2011} is a
collection of images depicting digits which were obtained from google
street view images. The dataset comes in two variants of which we
restrict ourselves to the one containing cropped $32 \times 32$ pixel
images. Similar to the well known MNIST dataset \cite{LeCun1998} the
task for this dataset is to classify each image as one of 10 digits in
the range from 0 to 9. The task is considerably more difficult than
MNIST since the images are cropped out of natural image data. The
images thus contain color information and show significant contrast
variation. Furthermore, although centered on one digit, several images
contain multiple visible digits, complicating the classification task.

\begin{table}[t]
\vskip 0.15in
\caption{Classification error of different models on the
  CIFAR-100 dataset.}
\begin{center}
\begin{small}
\begin{sc}
\begin{tabular}{l|l}
Method  & Error \\
\hline
Receptive Field Learning \cite{Jia2012} & 45.17 $\%$ \\
Learned Pooling \cite{Malinowski2013} & 43.71 $\%$ \\
Conv. Net + Stochastic Pooling \cite{ZeilerStochastic2013} & 42.51 $\%$ \\
Conv. Net + dropout + tree \cite{Nitish2013}  & \textbf{36.85 $\%$} \\
Conv. Net + Maxout  \cite{Goodfellow2013}  & 38.57 $\%$ \\
Conv. Net + Probout  & 38.14 $\%$ \\
\hline
\end{tabular}
\end{sc}
\end{small}
\end{center}
\vskip -0.1in
\label{cifar100_results}
\end{table}
The training and test set contain $73,257$ and $20,032$ labeled
examples respectively. In addition to this data there is an ``extra''
set of $531,131$ labeled digits which are somewhat less difficult to
differentiate and can be used as additional training data. As in
\cite{Goodfellow2013} we build a validation set by
selecting 400 examples per class from the training and 200 examples
per class from the extra dataset. We conflate all remaining training
images to a large set of $598,388$ images which we use for
training. 

The model trained for this task consists of three convolutional layers
containing 64, 128 and 128 units respectively, pooling over a 2
dimensional subspace. These are followed by a fully connected and a
softmax layer of which the fully connected layer contains 400 units
pooling over a 5 dimensional subspace. 
This yields a classification
error of $2.39 \%$ (using 50 model evaluations at test-time), matching
the current state of the art for a model trained on SVHN without data
augmentation achieved by the maxout model ($2.47 \%$). A
comparison to other results can be found in Table
\ref{svhn_results}~. This includes the current best result 
with data augmentation which was obtained using a generalization of
dropout in conjunction with a large network containing rectified
linear units \cite{WanLi2013}.

\begin{table}[h]
\vspace{-0.1cm}
\caption{Classification error of different models on the
  SVHN dataset. The top half shows a comparison of our result with the
  current state of the art achieved without data
  augmentation. The bottom half gives the best performance achieved
  with data augmentation as additional reference.}
\begin{center}
\begin{small}
\begin{sc}
\begin{tabular}{l|l}
Method  & Error \\
\hline
Conv. Net + Stochastic Pooling \cite{ZeilerStochastic2013} & 2.80 $\%$ \\
Conv. Net + dropout \cite{Nitish2013Mas} & 2.78 $\%$ \\
Conv. Net + Maxout \cite{Goodfellow2013}  & 2.47 $\%$ \\
Conv. Net + Probout  & \textbf{2.39 $\%$} \\
\hline
Conv. Net + dropout \cite{Nitish2013Mas} & 2.68 $\%$  \\
5 $\times$ Conv. Net + dropconnect \cite{WanLi2013} & \textbf{1.93 $\%$}  \\
\hline
\end{tabular}
\end{sc}
\end{small}
\end{center}
\vskip -0.1in
\label{svhn_results}
\end{table}

\section{Conclusion}
We presented a probabilistic version of the recently introduced maxout
unit. A model built using these units was shown to yield competitive
performance on three challenging datasets (CIFAR-10, CIFAR-100,
SVHN). As it stands, replacing maxout units with probout units is
computationally expensive at test time. This problem could be
diminished by developing an approximate inference scheme similar to
\cite{Krizhevsky2012, Zeiler2013} which we see as an interesting
possibility for future work. 

We see our approach as part of a larger body of work on exploring the
utility of learning ``complex cell like'' units which can give rise to
interesting invariances in neural networks. While this paradigm has
extensively been studied in unsupervised learning it is less explored
in the supervised scenario. We believe that work towards building
activation functions incorporating such invariance properties, while
at the same time designed for use with efficient model averaging
techniques such as dropout, is a worthwhile endeavor for advancing the
field.

\subsubsection*{Acknowledgments}
The authors want to thank Alexey Dosovistkiy for helpful discussions
and comments, as well as Thomas Brox for generously providing additional computing
resources.

{\footnotesize
\bibliography{papers.bib}

\begin{thebibliography}{10}

\bibitem{Hinton2012}
Alex Krizhevsky Ilya Sutskever Ruslan R.~Salakhutdinov Geoffrey E.~Hinton,
  Nitish~Srivastava.
\newblock Improving neural networks by preventing co-adaptation of feature
  detectors.
\newblock arxiv:cs/1207.0580v3.

\bibitem{Krizhevsky2012}
Alex Krizhevsky, Ilya Sutskever, and Geoff Hinton.
\newblock Imagenet classification with deep convolutional neural networks.
\newblock In {\em Advances in Neural Information Processing Systems 25}. 2012.

\bibitem{Zeiler2013}
Matthew Zeiler and Rob Fergus.
\newblock Visualizing and understanding convolutional networks.
\newblock arxiv:cs/1311.2901v3.

\bibitem{WanLi2013}
Li~Wan, Matthew~D. Zeiler, Sixin Zhang, Yann LeCun, and Rob Fergus.
\newblock Regularization of neural networks using dropconnect.
\newblock In {\em International Conference on Machine Learning (ICML)}, 2013.

\bibitem{Frey2013}
Jimmy Ba and Brendan Frey.
\newblock Adaptive dropout for training deep neural networks.
\newblock In {\em Advances in Neural Information Processing Systems 26}. 2013.

\bibitem{ZeilerStochastic2013}
Matthew~D. Zeiler and Rob Fergus.
\newblock Stochastic pooling for regularization of deep convolutional neural
  networks.
\newblock In {\em International Conference on Learning Representations (ICLR):
  Workshop track}, 2013.

\bibitem{Goodfellow2013}
Ian~J. Goodfellow, David Warde-Farley, Mehdi Mirza, Aaron Courville, and Yoshua
  Bengio.
\newblock Maxout networks.
\newblock In {\em International Conference on Machine Learning (ICML)}, 2013.

\bibitem{Nair2010}
Vinod Nair and Geoffrey~E. Hinton.
\newblock Rectified linear units improve restricted boltzmann machines.
\newblock In {\em International Conference on Machine Learning (ICML)}, 2010.

\bibitem{Glorot2011}
Xavier Glorot, Antoine Bordes, and Yoshua Bengio.
\newblock Deep sparse rectifier neural networks.
\newblock In {\em AISTATS 2011}, April 2011.

\bibitem{Hyvarinnen2009}
Jarmo~Hurri Aapo~Hyvärinen and Patrik~O. Hoyer.
\newblock {\em Natural Image Statistics}.

\bibitem{Bergstra2009}
Yoshua Bengio and James~S. Bergstra.
\newblock Slow, decorrelated features for pretraining complex cell-like
  networks.
\newblock In {\em Advances in Neural Information Processing Systems 22}. 2009.

\bibitem{Zou2012}
Will~Y. Zou, Shenghuo Zhu, Andrew~Y. Ng, and Kai Yu.
\newblock Deep learning of invariant features via simulated fixations in video.
\newblock In {\em Neural Information Processing Systems (NIPS 2012)}, 2012.

\bibitem{Gulcere2013}
Razvan Pascanu Yoshua~Bengio Caglar~Gulcehre, Kyunghyun~Cho.
\newblock Learned-norm pooling for deep neural networks.
\newblock arxiv:stat/1311.1780v3.

\bibitem{Lee2009}
Honglak Lee, Roger Grosse, Rajesh Ranganath, and Andrew~Y Ng.
\newblock {Convolutional deep belief networks for scalable unsupervised
  learning of hierarchical representations}.
\newblock pages 1--8, 2009.

\bibitem{BengioStochastic2013}
Yoshua Bengio.
\newblock Estimating or propagating gradients through stochastic neurons.

\bibitem{BengioGSN2013}
Jason~Yosinski Yoshua~Bengio, Éric Thibodeau-Laufer.
\newblock Deep generative stochastic networks trainable by backprop.

\bibitem{GoodfellowPylearn}
Pascal Lamblin Vincent Dumoulin Mehdi Mirza Razvan Pascanu James Bergstra
  Frédéric Bastien Yoshua~Bengio Ian J.~Goodfellow, David Warde-Farley.
\newblock Pylearn2: a machine learning research library.
\newblock arxiv:stat/1308.4214.

\bibitem{Krizhevsky2009}
A.~Krizhevsky and G.~Hinton.
\newblock Learning multiple layers of features from tiny images.
\newblock 2009.

\bibitem{Koray_CVPR2009}
Koray Kavukcuoglu, Marc'Aurelio Ranzato, Rob Fergus, and Yann LeCun.
\newblock Learning invariant features through topographic filter maps.
\newblock In {\em Proc. International Conference on Computer Vision and Pattern
  Recognition (CVPR)}, 2009.

\bibitem{Snoek2012}
Jasper Snoek, Hugo Larochelle, and Ryan~Prescott Adams.
\newblock Practical bayesian optimization of machine learning algorithms.
\newblock In {\em Advances in Neural Information Processing Systems 25},
  12/2012 2012.

\bibitem{Nitish2013}
Nitish Srivastava and Ruslan Salakhutdinov.
\newblock Discriminative transfer learning with tree-based priors.
\newblock In {\em Advances in Neural Information Processing Systems 26}, pages
  2094--2102. 2013.

\bibitem{Netzer2011}
Yuval Netzer, Tao Wang, Adam Coates, Alessandro Bissacco, Bo~Wu, and Andrew~Y.
  Ng.
\newblock Reading digits in natural images with unsupervised feature learning.
\newblock In {\em NIPS Workshop on Deep Learning and Unsupervised Feature
  Learning 2011}, 2011.

\bibitem{LeCun1998}
Yann LeCun, Léon Bottou, Yoshua Bengio, and Patrick Haffner.
\newblock Gradient-based learning applied to document recognition.
\newblock In {\em Proceedings of the IEEE}, number~11, 1998.

\bibitem{Jia2012}
Yangqing Jia, Chang Huang, and Trevor Darrell.
\newblock Beyond spatial pyramids: Receptive field learning for pooled image
  features.
\newblock In {\em CVPR}, 2012.

\bibitem{Malinowski2013}
Mateusz Malinowski and Mario Fritz.
\newblock Learnable pooling regions for image classification.
\newblock In {\em International Conference on Learning Representations (ICLR):
  Workshop track}, 2013.

\bibitem{Nitish2013Mas}
Nitish Srivastava.
\newblock Improving neural networks with dropout.
\newblock In {\em Master's thesis, University of Toronto, 2013.}

\end{thebibliography}
\bibliographystyle{hunsrt}}

\end{document}